\documentclass[a4paper,10pt]{article}

\usepackage{fourier}
\usepackage{color}
 \usepackage{graphicx}
\usepackage{url}
\usepackage[affil-it]{authblk}
\usepackage{amsmath}
\usepackage{wrapfig}
\usepackage[hidelinks]{hyperref}

\usepackage[T1]{fontenc}
\usepackage{times}

\begin{document}

\title{Towards a performance analysis on pre-trained Visual Question Answering models for autonomous driving}
\author{Kaavya Rekanar$^1$, Ciarán Eising$^1$, Ganesh Sistu$^2$, Martin Hayes$^1$}

\affil{{\textit{$^1$University of Limerick, $^2$Valeo Vision Systems}\\
$^1$firstname.lastname@ul.ie,
$^2$firstname.lastname@valeo.com}}
\date{}
\maketitle
\thispagestyle{empty}

\begin{abstract}

This short paper presents a preliminary analysis of three popular Visual Question Answering (VQA) models, namely ViLBERT, ViLT, and LXMERT, in the context of answering questions relating to driving scenarios. The performance of these models is evaluated by comparing the similarity of responses to reference answers provided by computer vision experts. Model selection is predicated on the analysis of transformer utilization in multimodal architectures.  The results indicate that models incorporating cross-modal attention and late fusion techniques exhibit promising potential for generating improved answers within a driving perspective. This initial analysis serves as a launchpad for a forthcoming comprehensive comparative study involving nine VQA models and  sets the scene for further investigations into the effectiveness of VQA model queries in self-driving scenarios. Supplementary material is available on the \href{https://github.com/KaavyaRekanar/Towards-a-performance-analysis-on-pre-trained-VQA-models-for-autonomous-driving}{Github} page. 

 \end{abstract}

\section{Introduction}

Visual Question Answering (VQA) is the process of generating natural language responses to open-ended questions by leveraging visual information derived from an image. This task encompasses the generation of textual answers to queries expressed in natural language. Visual question answering (VQA) holds significant importance for self-driving cars due to the requirement for enhanced perception and decision-making in autonomous vehicles. By incorporating VQA systems into the framework of self-driving cars, key benefits like Contextual Understanding, Enhanced Human-Machine Interaction, Adaptive Decision-Making, and Safety and Error Handling can be realized \cite{xiong2022challenges}. By integrating VQA capabilities, autonomous vehicles can enhance their perception, communication, and decision-making processes, ultimately leading to safer and more efficient driving experiences.

This paper  provides an introductory overview of the analysis conducted on three select models, focusing specifically on their performance in the domain of Visual Question Answering (VQA) with a strict focus on driving scenarios. It is part of a research study aimed at identifying the most effective VQA model for answering questions related to driving. Although there are review papers available on VQA models 
\cite{zhong2022video}, there is a notable research gap, as none of these studies has conducted a model evaluation in the context of common driving scenarios. A survey has been conducted to observe how pretrained available models respond to questions and how similar or different the answers are when compared to humans. The comparative analysis done has led us to the result that the available models are not as suitable for questions in a driving context as they are in a general scenario and this is a research gap that could be exploited. Additionally, the authors observed that there has not been a thorough performance analysis conducted on this topic. 

\section{Background Study and Related Work}

VQA models incorporate multimodal architectures that utilize transformers to handle the fusion of visual and textual modalities. Transformers enable contextual understanding and information exchange between the visual and textual components of the input, facilitating more accurate and comprehensive question answering \cite{vaswani2017attention}. Therefore, multimodal models employ transformers to process and fuse information from different modalities. Within the domain of VQA, multimodal models leverage transformers to handle the integration of visual and textual information, allowing for enhanced understanding and improved performance in answering questions based on visual inputs.

Transformers in multimodal models with vision and NLP refer to the application of transformer-based architectures in tasks that involve both visual and textual information. Transformers have demonstrated great success in natural language processing (NLP) tasks, thanks to their ability to capture long-range dependencies and model sequential data effectively. However, the integration of visual information poses unique challenges, and incorporating it into transformers allows for more powerful multimodal models.

Traditionally, multimodal models combined visual and textual information using separate pathways, such as using convolutional neural networks (CNNs) for image processing and recurrent neural networks (RNNs) for language processing. Transformers offer an alternative approach that enables joint processing of both modalities within a unified architecture.

Transformers are utilized in multimodal models for early fusion, late fusion, and cross-modal attention according to \cite{nagrani2021attention}. Early fusion involves the simultaneous processing of modalities to learn joint representations. Late fusion includes separate processing of modalities, followed by fusion to capture interdependencies. Cross-modal attention enables information exchange and alignment between modalities, enhancing multimodal understanding and integration. More details on how transformers are utilized in each of these types of models can be read in \cite{boulahia2021early} and \cite{nagrani2021attention}. 

\section{Methodology}
In this study, we collected a comprehensive corpus of 78 research papers on Visual Question Answering (VQA)\footnote{Full list of papers available at our  \href{https://github.com/KaavyaRekanar/Towards-a-performance-analysis-on-pre-trained-VQA-models-for-autonomous-driving}{Github page.}} From this collection, we carefully selected nine models based on specific criteria for our analysis. These models were evaluated for user interface quality, code replication ease, and compatibility with our pretrained models. The initial experiment aimed to enhance the models' performance using the German Traffic Sign Recognition Benchmark (GTSRB) dataset, explicitly focusing on signboard interpretation \cite{stallkamp2011german}. However, the results revealed limited comprehension of driving-related matters by the models. This led us to conduct an additional experiment with computer vision experts, presenting them with contextually minimal images from our dataset, mirroring the approach used with the pre-trained models.

For our experiment comparing human responses to multimodal models, we selected three models solely based on their utilization of transformers in their architectures from the previously mentioned nine models. A brief introduction about the three models chosen for the analysis:
\begin{itemize}
    \item Vision and Language BERT (ViLBERT)- Early Fusion: extends BERT with a co-attention mechanism, integrating vision-attended language features into visual representations. It enables joint reasoning about text and images for visual grounding \cite{lu2019vilbert}.

  \item Vision-and-Language Transformer (ViLT)- Cross-Modal Attention: aligns visual and textual features and generates joint representations through a visual encoder and a language encoder \cite{kim2021vilt}.

 \item Learning Cross-Modality Encoder Representations from Transformers (LXMERT)- Late Fusion: incorporates multi-level interactions between vision and language by employing cross-attention mechanisms. It captures the interplay between different modalities and generates accurate answers to the posed questions \cite{tan2019lxmert}.
\end{itemize}
The authors conducted a survey consisting of two specific questions, namely "What are the contents of the image?" and "What should the driver do?", targeting a carefully chosen set of images all pertaining to driving scenarios. These images were selected from the MS COCO dataset. The survey was distributed among a cohort of ten Computer Vision Experts who provided responses to the questions based on the available options and the accompanying images. The answer that received the most votes was selected as the ground truth. The comprehensive outcomes of this survey are presented in Figure 1.

The rationale behind asking both subjective and objective questions, namely "What are the contents of the image?" and "What should the driver do?", is to assess the model's ability to comprehend and respond to different types of questions in the context of visual information.
Subjective questions, like "What are the contents of the image?", require the model to understand and interpret the visual content and provide a descriptive answer. These questions evaluate the model's capability to recognize objects, scenes, and other relevant visual elements depicted in the image. Objective questions, like "What should the driver do?", require the model to provide a specific action or response based on the given visual information. These questions assess the model's understanding of driving scenarios and ability to reason about the appropriate course of action.

By including both subjective and objective questions, the experiment aims to evaluate different aspects of the model's performance. Subjective questions focus on the model's visual comprehension and scene understanding abilities, while objective questions assess its ability to provide contextually appropriate and practical responses in a driving context. This comprehensive evaluation helps to gauge the model's overall proficiency in visual question answering and its potential utility in real-world applications such as self-driving cars.

The ground truth for the respective questions was evaluated against the answers generated by the pre-trained models. Figure 1 provides a visual representation of the model's performance in addressing the posed questions, allowing for an assessment of their effectiveness based on the ground truth. The rationale behind comparing the answers of three Visual Question Answering (VQA) models with human answers and using colour coding (green for correct, orange for wrong, yellow for partially correct) is to visually highlight the performance and discrepancies between the models and human responses as done in \cite{dzelzkaleja2020color}. This visual representation allows for a quick and intuitive understanding of the accuracy and effectiveness of the models in comparison to human performance.

\section{Results and Analysis}
\begin{figure}[h!]
    \centering
    \includegraphics[scale=0.3]{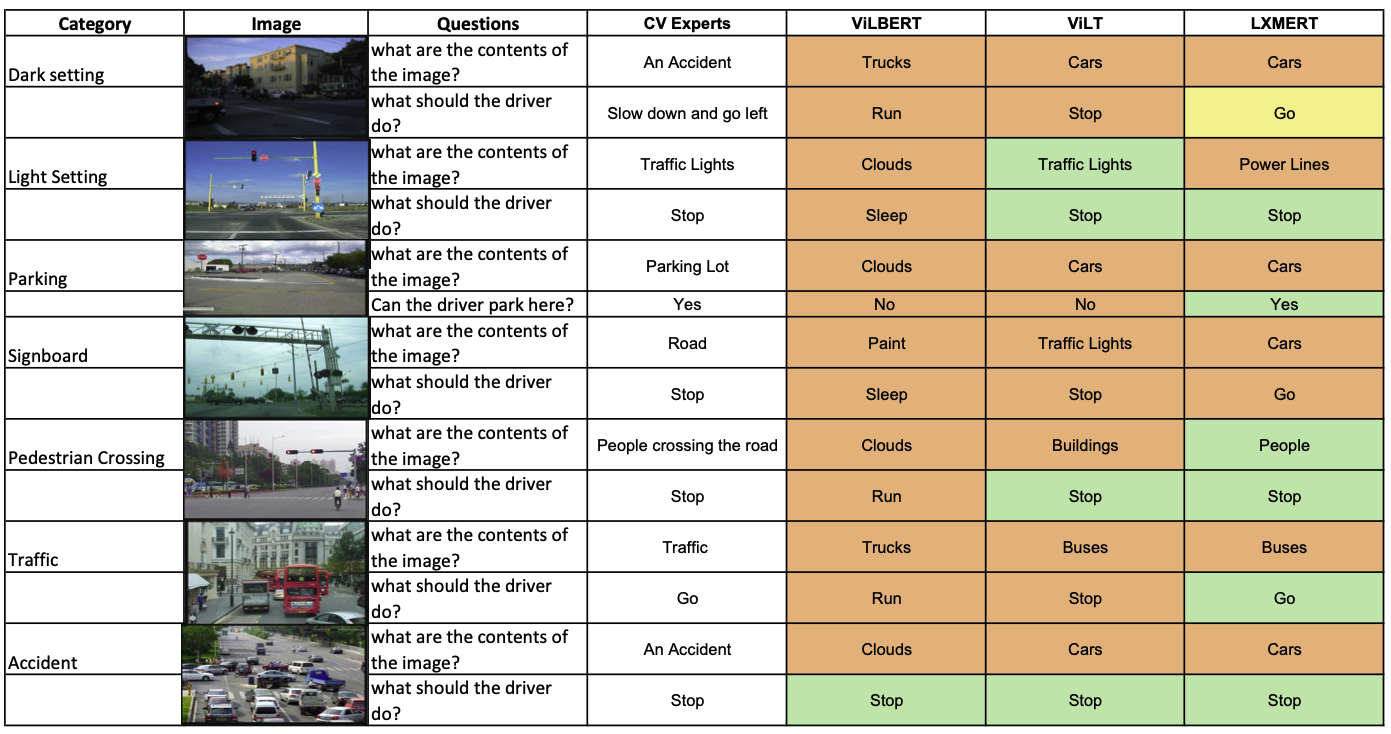}
    \caption{Comparison of Responses: Computer Vision Experts vs. Selected Models}
    \label{fig:results}
\end{figure}

Figure \ref{fig:results} concisely summarises the results from the experiment conducted on the selected models.

\subsection{Analysis}
In summary, the analysis of the evaluated models to date yields the following observations:
ViLBERT demonstrates a lack of comprehension regarding the question "What are the contents of the image?", as it consistently provided the answer ”nothing”. However, when posed with the question "What are the objects of the image?", ViLBERT manages to produce answers, albeit quite often incorrect or of limited utility for the application at hand. Consequently, ViLBERT is not an optimal choice for fine-tuning within the context of self-driving scenarios.
ViLT exhibits a certain level of capability in generating answers based on the provided images. Notably, when addressing the question "What should the driver do?", ViLT frequently responds with "Stop," as depicted in Figure~\ref{fig:results}. However, upon further investigation, it becomes apparent that the model does perform well in terms of question comprehension, and its object identification performance surpasses that of ViLBERT. This finding suggests that ViLT holds promise for fine-tuning with the GTSRB dataset, enabling it to learn how to effectively answer questions within a driving context.
The LXMERT model demonstrates better performance in answering questions within a driving context. Although the model exhibits excellent object identification capabilities, the accuracy of its answers requires refinement. The authors noted that LXMERT's object identification algorithm effectively recognizes objects in various scenes, and can offer accurate scene descriptions with the important exception of accident-related images. This observation implies there exists potential to enhancing LXMERT's performance in driving scenarios through fine-tuning with the GTSRB dataset, thereby improving its performance in driving-specific use cases.

\section{Conclusions and Future Work}
This paper has reviewed the performance of three VQA models, namely ViLBERT, ViLT, and LXMERT, from a driver assistance perspective, focusing on model efficacy in terms of similarity to expert responses for posed questions. Based on the analysis presented in this paper, it is inferred that both ViLT and LXMERT exhibit promising performance in this application space. 
However, despite the advancements observed in these models, further research and development are required to address the specific challenges associated with driver assistance. The ability to accurately comprehend and respond to user queries in real-time scenarios remains a crucial aspect of enhancing the interaction between drivers and vehicles. Achieving a VQA model that can effectively interpret diverse driver inquiries, provide accurate answers, and adapt to dynamic driving conditions is essential for optimizing user-car interaction.

Moving forward, the work will expand its scope by conducting a more comprehensive performance analysis that considers six additional selected models, including basic and fine-tuned pretrained models using the GTSRB dataset. The ultimate objective is to identify a preferred model that can be extensively trained with an expanded dataset that encompasses driving scenarios with subjective reference responses. Future research will focus on providing better codified contextual information to both experts and models, including camera location, velocity, acceleration, handbrake, and steering inputs, to enable more informed assessment of performance and to enable decisions on the next highest priority action to be taken with greater confidence. 

\section*{Acknowledgments}

This publication has emanated from research conducted with the financial support of Science Foundation Ireland under Grant number 18/CRT/6049. For the purpose of Open Access, the author has applied a CC BY public copyright licence to any Author Accepted Manuscript version arising from this submission.


\bibliographystyle{apalike}
\bibliography{references.bib}

\begin{thebibliography}{}

\bibitem[Boulahia et~al., 2021]{boulahia2021early}
Boulahia, S.~Y., Amamra, A., Madi, M.~R., and Daikh, S. (2021).
\newblock Early, intermediate and late fusion strategies for robust deep
  learning-based multimodal action recognition.
\newblock {\em Machine Vision and Applications}, 32(6):121.

\bibitem[Dzelzkaleja, 2020]{dzelzkaleja2020color}
Dzelzkaleja, L. (2020).
\newblock Color code method design evaluation and data analysis.
\newblock {\em International Journal of Engineering \& Technology},
  7(2.28):106--109.

\bibitem[Kim et~al., 2021]{kim2021vilt}
Kim, W., Son, B., and Kim, I. (2021).
\newblock Vilt: Vision-and-language transformer without convolution or region
  supervision.
\newblock In {\em International Conference on Machine Learning}, pages
  5583--5594. PMLR.

\bibitem[Lu et~al., 2019]{lu2019vilbert}
Lu, J., Batra, D., Parikh, D., and Lee, S. (2019).
\newblock Vilbert: Pretraining task-agnostic visiolinguistic representations
  for vision-and-language tasks.
\newblock {\em Advances in neural information processing systems}, 32.

\bibitem[Nagrani et~al., 2021]{nagrani2021attention}
Nagrani, A., Yang, S., Arnab, A., Jansen, A., Schmid, C., and Sun, C. (2021).
\newblock Attention bottlenecks for multimodal fusion.
\newblock {\em Advances in Neural Information Processing Systems},
  34:14200--14213.

\bibitem[Stallkamp et~al., 2011]{stallkamp2011german}
Stallkamp, J., Schlipsing, M., Salmen, J., and Igel, C. (2011).
\newblock The german traffic sign recognition benchmark: a multi-class
  classification competition.
\newblock In {\em The 2011 international joint conference on neural networks},
  pages 1453--1460. IEEE.

\bibitem[Tan and Bansal, 2019]{tan2019lxmert}
Tan, H. and Bansal, M. (2019).
\newblock Lxmert: Learning cross-modality encoder representations from
  transformers.
\newblock {\em arXiv preprint arXiv:1908.07490}.

\bibitem[Vaswani et~al., 2017]{vaswani2017attention}
Vaswani, A., Shazeer, N., Parmar, N., Uszkoreit, J., Jones, L., Gomez, A.~N.,
  Kaiser, {\L}., and Polosukhin, I. (2017).
\newblock Attention is all you need.
\newblock {\em Advances in neural information processing systems}, 30.

\bibitem[Xiong et~al., 2022]{xiong2022challenges}
Xiong, W., Fan, H., Ma, L., and Wang, C. (2022).
\newblock Challenges of human—machine collaboration in risky decision-making.
\newblock {\em Frontiers of Engineering Management}, 9(1):89--103.

\bibitem[Zhong et~al., 2022]{zhong2022video}
Zhong, Y., Ji, W., Xiao, J., Li, Y., Deng, W., and Chua, T.-S. (2022).
\newblock Video question answering: datasets, algorithms and challenges.
\newblock {\em arXiv preprint arXiv:2203.01225}.

\end{thebibliography}

\end{document}